\documentclass[12pt]{article}

\usepackage[a4paper, left=3cm, top=3cm, right=3cm, bottom=3cm]{geometry}
\usepackage{authblk}
\usepackage{amsmath, amsfonts, amssymb, amsthm}
\usepackage{graphicx}
\usepackage{booktabs}
\usepackage{mathtools}
\usepackage{enumitem}
\usepackage{algorithm}
\usepackage{algpseudocode}

\usepackage{tocloft}

\newcommand{\R}{\mathbb{R}}
\newcommand{\N}{\mathbb{N}}

\newcommand{\Ao}{\mathbf{A}}
\newcommand{\Mo}{\mathbf{M}}

\newcommand{\csdata}{Y}
\newcommand{\pdata}{P}
\newcommand{\source}{u}

\newcommand{\rr}{\mathbf r}
\newcommand{\rss}{\mathbf s}

\newcommand{\spin}{\Theta}

\cftpagenumbersoff{figure}
\cftpagenumbersoff{table}

\usepackage[binary-units]{siunitx}
\DeclareSIUnit{\sample}{S}

\newcommand{\x}{\mathbf{x}}
\newcommand{\z}{\mathbf z}
\newcommand{\y}{\mathbf y}

\newcommand{\signal}{\mathbf{x}}

\newcommand{\Wo}{\mathbf{W}}
\DeclareMathOperator{\To}{\boldsymbol{\Phi}}

\DeclarePairedDelimiter{\norm}{\lVert}{\rVert}

\newcommand\abs[1]{\left\vert#1\right\vert}
\newcommand\set[1]{\left\{#1\right\}}
\newcommand\sabs[1]{\lvert#1\rvert}
\newcommand{\kl}[1]{\left(#1\right)}
\newcommand{\ekl}[1]{[#1]}

\newcommand{\rmd}{\mathrm d}

\newcommand{\B}{\mathbf B}

\newcommand{\argmin}{\mathrm{argmin}}

\newtheorem{theorem}{Theorem}[section]\newtheorem{definition}[theorem]{Definition}
\newtheorem{remark}[theorem]{Remark}

\usepackage{setspace}
\usepackage{multirow}

\newcommand{\blau}{}

\title{Design, Implementation and Analysis of a Compressed Sensing Photoacoustic Projection Imaging System}

\title{Design, Implementation and Analysis of a Compressed Sensing Photoacoustic Projection Imaging System}

\date{February 24, 2024}

\author[a,*]{Markus Haltmeier}
\author[a]{Matthias Ye}

\affil{Department of Mathematics, University of Innsbruck, 6020 Innsbruck, Austria,
E-mail:  \texttt{markus.haltmeier@uibk.ac.at}
 }

\author[b]{Karoline Felbermayer}
\author[b]{Florian Hinterleitner}
\author[b]{Peter Burgholzer}

\affil[b]{Research Center for Non Destructive Testing (RECENDT), 4040 Linz, Austria}

\begin{document}

\maketitle

\begin{abstract} \mbox{}

\medskip\noindent\textbf{Significance:}
Compressed sensing (CS) uses special measurement designs combined with powerful mathematical algorithms to reduce the amount of data to be collected while maintaining image quality. This is relevant to almost any imaging modality, and in this paper we focus on CS in photoacoustic projection imaging (PAPI) with integrating line detectors (ILDs).

\medskip\noindent\textbf{Aim:}
Our previous research involved rather general CS measurements, where each ILD can contribute to any measurement. In the real world, however, the design of CS measurements is subject to practical constraints. In this research, we aim at a CS-PAPI system where each measurement involves only a subset of ILDs, and which can be implemented in a cost-effective manner.

\medskip\noindent\textbf{Approach:}
We extend the existing PAPI with a self-developed CS unit. The system provides structured CS matrices for which the existing recovery theory cannot be applied directly. A random search strategy is applied to select the CS measurement matrix within this class for which we obtain exact sparse recovery.     

\medskip\noindent\textbf{Results:}
We implement a CS PAPI system for a compression factor of $4:3$, where specific measurements are made on separate groups of 16 ILDs. We algorithmically design optimal CS measurements that have proven sparse CS capabilities.  Numerical experiments are used to support our results.

\medskip\noindent\textbf{Conclusions:}  
CS with proven sparse recovery capabilities can be integrated into PAPI, and numerical results support this setup. Future work will focus on applying it to experimental data and utilizing data-driven approaches to enhance the compression factor and generalize the signal class.


\end{abstract}

\section{Introduction}
\label{sec:intro}

Photoacoustic tomography (PAT) is an emerging non-invasive imaging technique that combines the high contrast of optical imaging with the high spatial resolution of ultrasound imaging \cite{wang2006photoacoustic, wang2015photoacoustic, wang2012photoacoustic}. It is based on the generation of acoustic waves by illuminating a sample with picosecond or nanosecond optical pulses.  The acoustic waves are measured outside the object and mathematical algorithms are used to reconstruct an image of the inside.  While there are many important practical and theoretical aspects along the lines of signal generation, signal detection, system design, image generation and enhancement, in this paper we focus on the measurement and inversion of acoustic waves \cite{poudel2019survey, rosenthal2013acoustic}.  Specifically, we focus on PA projection imaging (PAPI) based on integrating line detectors (ILDs) \cite{burgholzer2005thermoacoustic, paltauf2017photoacoustic}. Our goal is to use ideas from compressed sensing (CS) to reduce the number of spatial measurements compared to standard measurements where each ILD  is used to record its own time-dependent signal. Specifically, we present our   design and development of CS in PAT under physical constraints that naturally arise in the already existing self-developed PAPI system \cite{bauer2017all}.

\subsection{Photoacoustic projection imaging (PAPI)}

A  PA projection tomograph records the induced acoustic signals with an array of parallel ILDs, with each sensor integrating (averaging) the pressure along the lines of the detectors. The data thus consists of  samples  of the linear projection of the 3D acoustic pressure wave  in the direction of the ILDs. Reconstruction in 2D gives a projection of the initial pressure distribution.  If a 3D reconstruction is required, the object can be rotated around an axis perpendicular to the fibers, and a 3D reconstruction is computed from the collection of 2D projections by inversion of the 2D Radon transform, which is similar to parallel beam X-ray CT \cite{paltauf2007photacoustic, burgholzer2007temporal}. As in X-ray imaging, where in certain situations single projections are sufficient, the same can be said for photoacoustic imaging. We will therefore  restrict ourselves to 2D PAPI.

\begin{figure}[htb!]
\centering \includegraphics[width=0.8\textwidth]{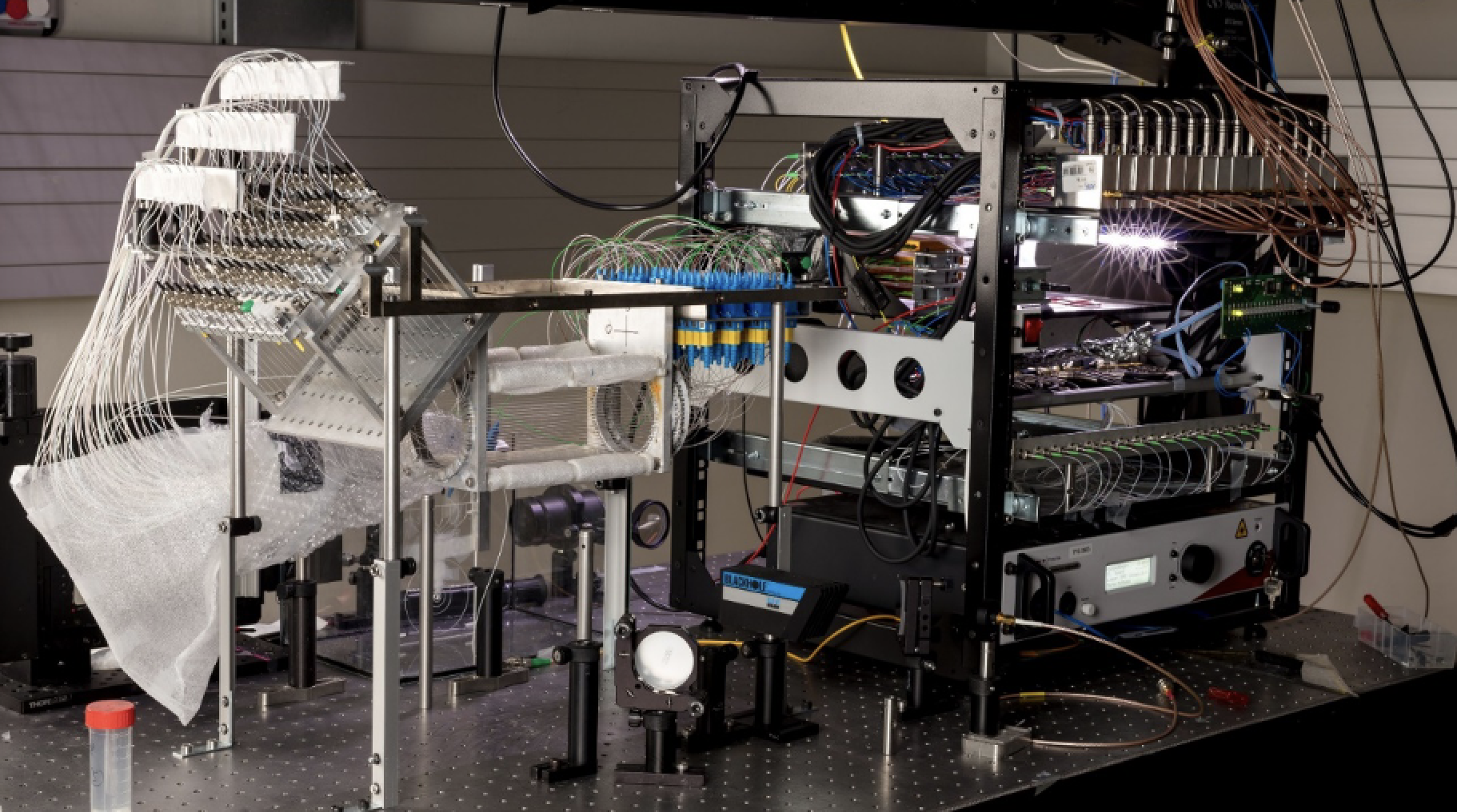}
\caption{Photographic image of the PA projection tomography with 64 fiber optic Mach-Zehnder interferometers (FOMZIs) as ILDs forming  the  basics  of the presented research.} \label{fig:papi}
\end{figure}

Figure~\ref{fig:papi} shows a photograph of our self-developed all-optical PAPI system used in this study. The setup is based on fiber optic Mach-Zehnder interferometers (FOMZIs) with graded index polymer optical fibers (GIPOFs). These have a higher bandwidth than glass optical fibers, are more stable for measurements.  In the current system, 64 ILDSa are arranged on a circle forming a cylinder. Readout for each sensor requires an analog-to-digital (AD) converter, and 4 sensors are multiplexed to one AD converter. Thus, to measure all 64 signals, the measurement process must be repeated four times.  Our hypothesis is that proper combinations of ILD signals will be advantageous over recording individual signals when used in conjunction with a nonlinear CS recovery algorithm.

\subsection{Compressed sensing (CS) in PAPI}

Following the CS paradigm, instead of recording pressure signals $P = [p_1^T, \dots, p_n^T]^T$ where $p_j$ is the pressure signal (written as column vector) of the $j$-th ILD, we record CS data
\begin{equation} \label{eq:cs-data}
	y_i = (\Ao \pdata)_i = \sum_{j=1}^n a_{ij} p_j \quad \text{ for } i \in \{1,2, \dots , m\} \,,
\end{equation}
with $\Ao = (a_{i,j})_{i,j} \in \R^{m \times n}$ denoting the CS measurement matrix. Usually in  CS, the measurement matrix chosen randomly, since this gives exact recovery of sparse vectors with a high probability  for large $n,m$. However, in practice, and specifically in our application, the matrix $\Ao$ cannot be chosen completely at random. First, the measurements cannot  combine all pressure values if they are not connected to the same controller. Second, the numbers $a_{i,j}$ are often restricted to specific values, in our case for example to 0 and 1. Finally, the dimensionality $n$ in  our case is small, which limits the applicability of existing asymptotic CS theory that applies to the limit $n,m \to \infty$.   

The goal of this work is to design, analyze, and implement a CS  strategy that can actually be realized with  our PAPI system.  Within the  considered family of measurements, we investigate the optimal design of matrices. Due to the  low dimensionality of CS matrices,  even a small compression factor $n/m$  below 2 seems to be a substantial challenge.

\subsection{Outline}

In this paper, we present our finding and results in building a CS-PAPI system. This development is based on several steps. First, we provide a rigorous description of the PAPI problem. In this context, we also provide an overview of the most important background knowledge required. Second, we  introduce a novel class of CS measurements that are practically feasible and can be realized  with the existing self-developed PAPI setup. Third, we present a concept of optimal measurement design that allows researchers and practitioners to strategically select measurements to maximize imaging accuracy for CS in PAPI and other imaging modalities.  While  these results are developed in the context of sparsity we presents an  outlook for the use of more general signal classes potentially enabling data driven machine learning methods. Finally, we go from  theory to practice and show how these results can be translated into the experimental realization of CS-PAPI.

\section{Background}

In this section we present the background of our work.  This includes PAPI modeling (subsection~\ref{sec:papi}), sparse CS theory (subsection~\ref{sec:cs}) and the description of the self-developed PAPI system (subsection \ref{sec:papi-linz}).

\begin{figure}[thb!]
\begin{center}
  \includegraphics[width=0.9\textwidth]{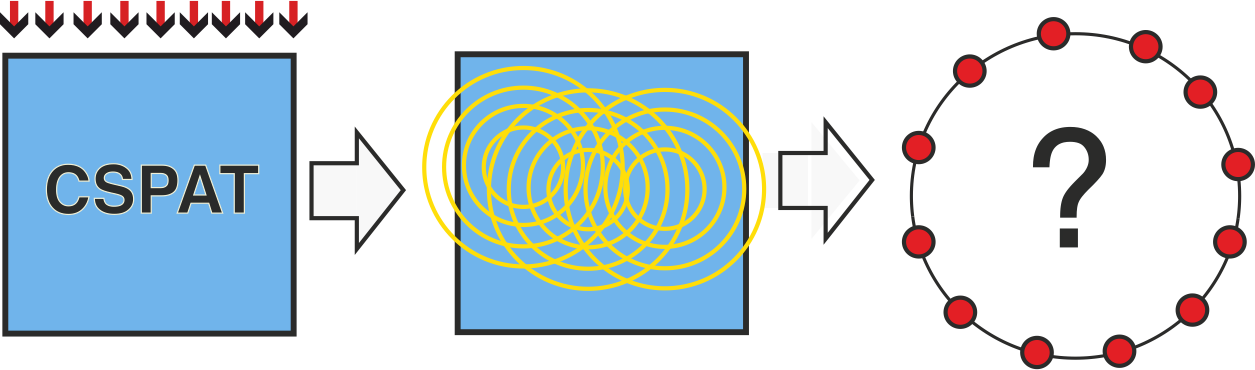}
\caption{(a) An object is illuminated with a  short optical pulse;  (b) the absorbed light distribution causes an acoustic pressure; (c) the  acoustic pressure is measured with ILD arranged  on a circle.\label{fig:pat}}
\end{center}
\end{figure}

\subsection{PA projection imaging (PAPI)}
\label{sec:papi}

PA tomography is based on generating an acoustic wave inside some investigated object using short optical pulses. When measuring  the pressure with ILDs, the imaging problem reduces to a 2D version of the standard problem \cite{paltauf2007photacoustic, burgholzer2007temporal} and in this  work we consider the  2D version only. Further, we restrict ourselves to constant sound speed and  a circular measurement geometry as illustrated in Figure~\ref{fig:pat}.

Let us denote by $\source \colon \R^2 \to \R$ the 2D PA source distribution which is our image of interest {\blau and} supposed to be enclosed by a circle $C_R$ of radius $R$. The 2D projected pressure satisfies the 2D wave equation
\begin{equation} \label{eq:wave}
\partial^2_t p (\rr,t)  - v_s^2 {\blau \Delta_{\rr}} p(\rr,t)
= \delta'(t)  \source (\rr)  \quad \text{ for } (\rr,t) \in \R^2 \times \R_+ \,,
\end{equation}
where $\delta'(t)$ is the  first time derivative of the Dirac delta distribution,  $\rr  \in \R^2$ is the spatial location, $t \in \R$ the time variable, $\Delta_{\rr}$ the spatial Laplacian and $v_s$ the constant speed of sound. The wave equation \eqref{eq:wave} is augmented with
$p(\rr, t) =0$  for $t < 0$ such that the acoustic pressure is uniquely defined  as solution of \eqref{eq:wave}. We rescale time in such a way that $v_s=1$.

PAPI in circular geometry consist in recovering the function $\source$ from measurements of $\Wo  \source (\rss,t) = p(\rss,t) $ made on $C_R  \times  (0,\infty)$. In the case of full data, exact  and  stable PA image reconstruction is  possible and several efficient methods for recovering $\source$   are available. We will use the FBP formula derived in \cite{FinHalRak07}
\begin{equation} \label{eq:fbp2d}
          \source(\rr)
         =
        - \frac{1}{\pi R}
        \int_{C_R}
        \int_{\abs{\rr-\rss}}^\infty
        \frac{ (\partial_t t \Wo  \source)(\rss, t)}{ \sqrt{t^2-\sabs{\rr-\rss}^2}}  \, \rmd t
        \rmd C(\rss)
         \,.
\end{equation}
Note the inversion operator in \eqref{eq:fbp2d} is also the adjoint of the forward operator $\Wo$. This in particular implies that inverting $\Wo$ is stable.

In practical applications, the acoustic pressure  can only be
measured with a finite number of acoustic detectors.
The standard sampling scheme  in a circular geometry  assumes
uniformly sampled values
\begin{equation} \label{eq:data}
    p \kl{ \rss_j, t_\ell}
    \text{ for }
    ( j, \ell) \in \set{ 1, \dots,  n} \times \set{ 1, \dots,  q }\,,
\end{equation}
with $\rss_j  \triangleq  R(  \cos \kl{\Omega (j-1)/n} , \sin \kl{\Omega (j-1)/n} )$, $t_\ell \triangleq 2R (\ell-1) /(q-1)$, and  $\Omega \leq 2\pi$ denoting  the angular covering on the detection circle. The number $n$ of detector positions in  \eqref{eq:data} is directly related to the resolution of the final reconstruction. Namely, $n \geq 2 R \lambda$ equally spaced transducers {\blau  covering the full circle} are required  to stably recover any PA source $\source$  that has  maximal essential wavelength  $\lambda$; see \cite{haltmeier2016sampling}. Image reconstruction in this case can be performed   by discretizing the inversion formula  \eqref{eq:fbp2d}. The sampling  condition  requires a very high sampling rate,  especially when the PA source contains narrow features, such as blood vessels or sharp interfaces.  {\blau  Commonly, $\lambda$ will be determined by the spatial sampling via the Nyquist condition, such that $2 R \lambda = \pi N_\rr$, where $N_\rr \times N_\rr$ is the number of samples for discretizing the object of interest on the square $[-R, R] \times [-R, R]$. In this case, we get $n = \operatorname{round}(\pi N_\rr / 2)$ for correct sampling according to Shannon Sampling theory.}

Note that temporal samples can easily be collected at a high sampling rate compared to the spatial sampling, where each sample requires a separate sensor. {\blau It is therefore beneficial to keep $n$ as small as possible by using tools that overcome the limitations of classical Shannon Sampling theory.}  Consequently, full sampling  is costly and time consuming and strategies for reducing the number of detector locations are desirable. In this study we use $n=64$ samples  which does  not satisfy the Nyqvist criteria for the targeted discretization. However the image quality in this case is  still reasonable.  To further  reduce  the  number of measurements while preserving  image  quality we use CS techniques.

\subsection{Compressed sensing (CS)}
\label{sec:cs}

The traditional approach to signal and image processing is to first collect a large number of point-like samples, which are then compressed and transmitted with minimal information loss. The basic idea of CS is to combine signal acquisition and compression by using specific indirect measurements together with mathematical algorithms that exploit inherent structure of the image. In this way, a high quality image can be recovered from a  smaller number of measurements than required for point sampling at the same resolution. In particular, the seminal works \cite{donoho2006compressed, candes2006robust} invented a theory of CS based on the sparsity of the signal to be recovered and the randomness of the measurements.  Subsequent research has identified properties of the measurement matrix, such as the restricted isometry property (RIP), as key elements for stable and robust recovery.

The first basic ingredient of  CS  is sparsity, that is defined as follows. Let $s \in \N$ and $\signal \in \R^n$. The  vector $\x$ is called $s$-sparse,  if  $\norm{\signal}_0 \coloneqq \sharp (\set{i  \in \set{1, \dots , n} \mid \signal\ekl{i} \neq 0})  \leq s$ where  $\sharp (S)$ stands for the number of elements in a set $S$. {\blau Signals} of practical  interest are often not sparse in the strict sense, but can be well approximated  by sparse vectors.One calls $ 	\sigma_{s} (\signal)	\coloneqq \inf \{ \norm{\signal - \signal_s}_1
	\mid  \signal_s \in \R^n \textnormal{ is $s$-sparse} \}$
the best  $s$-term approximation error of $\signal \in \R^n$, and  calls $\signal$ compressible, if  $\sigma_s(\x)$ decays sufficiently fast with increasing $s$.

\subsubsection{The restricted isometry constant (RIP)}

Let $s \in \N$ and $\delta \in (0,1)$.  Stable and robust recovery  of sparse vectors  requires the measurement matrix to well separate sparse vectors. The RIP  guarantees   such a separation. We recall that the  measurement matrix $\Ao \in \R^{m\times n}$ is said to satisfy the RIP of order $s$ with constant $\delta$ if
\begin{equation}\label{eq:rip}
(1-\delta)\norm{\signal}_2^2 \leq \norm{\Ao\signal}_2^2 \leq (1+\delta) \norm{\signal}_2^2
\quad \text{ for all $s$-sparse $\x \in \R^n$} \,,
\end{equation}
and  write $\delta_s$ for the smallest constant  satisfying (\ref{eq:rip}). Many sparse recovery results have been derived {\blau using} the RIP. For example, the result derived in \cite{CaiZha13} states that if  $\Ao \in \R^{m\times n}$ satisfies the $2s$-RIP with constant $\delta_{2s} < 1/2$ then  
 for  $\norm{ \y - \Ao\signal }_2 \leq \delta$ any $ \x_\star \in \argmin \{ \norm{\z}_1 \mid  \norm{\Ao \z - \y}_2 \}$ satisfies $\| \x - \x_\star \|_2 \leq   c_1  \sigma_{s}(\x) /  \sqrt{s} +  c_2 \delta$ for constants $c_1, c_2$ depending only on $\delta_{2s}$. This implies stable and robust recovery for measurement matrices satisfying the RIP. The error estimate consists of two terms: The term $c_2 \epsilon$ is due to the data noise and $c_1 \sigma_{s}(\mathbf{x}) / \sqrt{s}$ accounts for the fact that the unknown may not be strictly $s$-sparse.

No deterministic construction is known providing large measurement matrices satisfying the RIP with near-optimal $s$. However, several types of random matrices are known to satisfy the RIP with high probability. An important example of a random matrix that satisfies the RIP is the Bernoulli matrix, which is a random matrix $\B \in \set{-1,1}^{m\times n}$ having independent entries that take the values $-1$ and $1$ with equal probability. A Bernoulli matrix satisfies $\delta_{2s} < \delta$ with probability tending to $1$ as $m \to \infty$, provided that $ m \geq  C_\delta  s ( \log ( n/s) + 1 )$ for some constant $C_\delta > 0$ as $n \to \infty$. However, such a theory is hardly applicable in our situation due to the small dimension of our measurement matrices.

\subsubsection{Binary  CS matrices}

Another useful type of CS matrices are binary matrices having entries 0 or 1. Such measurement matrices can be interpreted as the adjacency matrix of a bipartite graph $(L, R, E)$ where $L \coloneqq {1, \dots, n}$ is the set of left vertices, $R \coloneqq \{1, \dots, m\}$ the set of right vertices, and $E \subseteq L \times R$ is the set of edges. Any element $(j, i) \in E$ can be interpreted as an edge joining vertices $j$ and $i$. The left vertices $L$ represent the sensors, and the right vertices $R$ model each measurement. The vertex $j \in L$ is connected to the vertex $i \in R$ if sensor $j$ contributes to measurement $i$. For our application, we have this type of binary measurement matrices.

Specific binary measurements are lossless expanders for which a stable and robust recovery theory exists \cite{BerGilIndKarTra08, FouRau13}.  However, these results are again of asymptotic nature and are not applicable for  PAPI  with small CS matrices.

\subsection{All-optical PA projection tomograph}
\label{sec:papi-linz}

In order to realize photoacoustic projection tomography one needs one or several ILDs that integrate the pressure along one dimension. Initial setups used a single line detector that is moved around the object either using a free-beam  Mach–Zehnder interferometer \cite{paltauf2007photoacoustic} or a free-beam as well as fiber-based  Fabry-Perot interferometer, \cite{burgholzer2006thermoacoustic}. To accelerate the data collection process arrays of line detectors have been developed either consisting of a piezoelectric array \cite{paltauf2017piezoelectric} or an array of FOMZIs introduced in  \cite{bauer2015photoacoustic, bauer2017all}. Optical and piezoelectric ILDs have been compared in \cite{nuster2019comparison}. A method where a PA projection image is collected at one shot is the full-field technique  \cite{nuster2010full}. In this paper we use the FOMZI array reviewed below.

\begin{figure}[htb!]
\includegraphics[width=\textwidth]{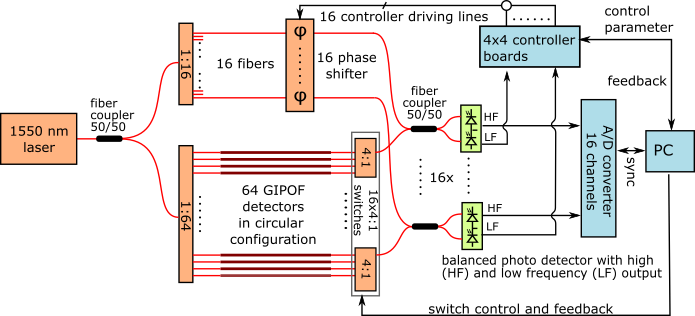}
\caption{Schematics of the photoacoustic projection imaging (PAPI) system setup using 64 detector positions.} \label{fig:fomzi}
\end{figure}

The PAPI setup consists of 18 individually designed (CAD) parts, for a total of 750 mechanical components. The fiber cage of the system is built with 64 GIPOFs, and each GIPOF has two end faces/ferrules, and five glue points, making a total of 128 end faces and 320 glue points. The fiber laser used is an NKT Koheras AdjustiK E15 with a maximum power of \SI{200}{\milli\watt} and a line width of 0.1kHz. A 1:2 fiber splitter directly after the fiber laser splits the optical path into a reference arm with \SI{20}{\percent} laser power and a measurement arm with \SI{80}{\percent} laser power. The 80/20 splitting is used because the measurement arm is split into 64 beams using a 1:64 fiber splitter whereas the reference arm is only split into 16 beams. Thus, each of the 80 fibers receives \SI{1.25}{\percent} of the overall laser power. The measurement arm consists of 64 GIPOFs arranged in a circular configuration and multiplexed with sixteen $4 \times 1$ fast fiber optic switches from Sercalo. The 16 fiber optic switches are controlled by the measurement software.

For working point stabilization of the FOMZIs, 16-fiber phase switches are integrated on 4 controller cards. A robust analog (bang-bang) controller with digital potentiometers and easy USB control was developed at RECENDT \cite{bauer2015photoacoustic}. The reference and measurement arms are connected by sixteen 2:2 50/50 fiber couplers and the 16 self-developed balanced photodetectors detect the optical signal and provide two electrical signals. A low-frequency (LF) signal is employed for working point stabilization, while a high-frequency (HF) signal represents the actual data.  The 16 PA  signals are sampled by a National Instrument (NI) device with 2  cards, each with 8 channels resulting in 16 channels in total. Each card has a maximum sampling rate of \SI{60}{\mega\sample\per\second}, \SI{12}{\bit} depth and \SI{128}{\mega\byte} on-board memory. The whole system is controlled by a PC with our own control and measurement software (NI LabWindows).

\section{System design, implementation and analysis}

In this section we present details on  the design, implementation and analysis  of our self-developed CS-PAPI device. It is build upon an extension  of the all-optical  PAPI described in Section~\ref{sec:papi-linz} using  specific  CS measurements that we optimize  by  introducing the sparse injectivity number (SIN) as a quality measure for CS measurement matrices.

\subsection{Compressive PAPI}
\label{sec:cs-PAT}

{\blau We conduct CS measurements of the pressure $P = \Wo \source$ in the detector domain, ensuring that pressures from different times are not mixed.} Thus, instead of  collecting $m$ individually sampled signals as in~\eqref{eq:data}, we take CS measurements $y_{i, \tau} = (\Ao \pdata)_{i,\tau}  \triangleq \sum_{k=1}^n	a_{i,j} p_{j , \ell} $ for $(i,\ell)  \in \set{  1, \dots, m }  \times  \set{ 1, \dots, q } $ with  $m < n$. Recall that $n$ is the number of sensors, $m$ the number of measurements and $q$ the number of temporal samples. If we write $\Wo \signal  = [(\Wo_1 \signal)^T ,  \dots,  (\Wo_n\signal)^T ]^T$ as a block column vector where the $j$-th row is the signal of the $j$-th  ILD, the CS-PAPI data can be written as  
\begin{equation} \label{eq:ip}
	\csdata =   
	\begin{bmatrix} 
	y_1   
	\\
	\vdots 
	\\
	y_m  
	\end{bmatrix}  
	= \Ao    
	\begin{bmatrix} 
	\Wo_1   \source 
	\\
	\vdots 
	\\
	\Wo_n   \source  
	\end{bmatrix} 
	=
	 \Ao    \Wo   \source  \,,
 \end{equation}
where the $y_i$ is $i$-th CS measurement signal. 

The aim of CS-PAPI image reconstruction is to recover the unknown $\source$ from data in  \eqref{eq:ip}.   If the matrix $\Ao$ would have Rank $n$, then \eqref{eq:ip} would  have the solution $\source  = \Wo^\sharp [ (\Ao^T \Ao)^{-1} \Ao^T \csdata ]$, where  $\Wo^\sharp$ is a numerical realization  of  the inversion formula  of the wave equation and  $(\Ao^T \Ao)^{-1} \Ao^T$  is the least square inverse of $\Ao$. In the case of compressive measurements, however, we have $m < n$ and the matrix $\Ao^T \Ao$ is singular. Thus solving $\csdata = \Ao \Wo \source$ becomes underdetermined and reconstruction algorithms using specific prior information are required. Following the prime CS strategy we use sparsity for that purpose.

Several choices for the CS measurement matrix $\Ao$ have been suggested  for PAT \cite{sandbichler2015novel, haltmeier2016compressed, betcke2016acoustic}. Specifically, for  PAPI with ILDs binary CS matrices are often most easily  realized in practice. In  this case sparsifying transformations in the detector domain may negatively affect stable recovery results. Note that the  CS measurement matrix $\Ao$ in \eqref{eq:ip} does not act in the temporal variable. Thus for any operation $\To$ acting in temporal variable only, we have the commutation relation {\blau $\Ao \circ \To = \To \circ \Ao$. This has been the motivation for the two-step image reconstruction approach proposed in \cite{sandbichler2015novel}, based on sparsifying temporal transforms, which we essentially follow here.} However, in contrast to that paper, we use a structured CS measurement matrix where only certain sensor combinations are allowed to be guided by the experimental design.

\subsection{Proposed structured CS measurements}

Recall that the PAPI  system (see Figure~\ref{fig:papi}) consists of 64 ILDs in total wich naturally come in 16 blocks of 4 sensors each, where each of these blocks is characterized  by sensors being connected to the same switch. We form CS measurements by selecting at most one sensor of each block and summing the signals over four neighboring blocks. In that way we make four CS  measurements in parallel where the  first measurement uses  detectors in group $[1] = \{1, \dots,  16\}$,  the second  in group $[2] = \{16, \dots,  32\}$,  the third in group  $[3] = \{33, \dots,  48\}$ and the fourth in group $[4] = \{49, \dots,  64\}$. In every measurement there is at  most one ILD active within one block  and every other sensor is inactive.  Making $m_0$ such  measurements, results for {\blau each group} in a binary $m_0 \times 16 $ matrix    
\begin{equation} \label{eq:submatrix}
\Ao_G  = \bigl[ \Ao_{G,1} \lvert \Ao_{G,2}  \vert \Ao_{G,3}  \vert  \Ao_{G,4} \bigr] \in \{0,1\}^{1 \times 16} \quad \text{ for } G=[1],[2],[3],[4] \,,
\end{equation}
where each block $\Ao_{G,b}$ has  at most one non-vanishing entry.   
Entry $1$ means the corresponding  sensor is active  and $0$ means that the sensor is inactive. An example  for such a matrix with $m_0 = 2$ measurements is 
\begin{equation*}
\Ao_G =
\begin{bmatrix}
     \begin{array}{cccc|cccc|cccc|cccc}
        1 & 0 & 0 & 0 & 0 & 0 & 0 & 0 & 1 & 0 & 0 & 0 & 0 & 1 & 0 & 0 \\ \hline
        0 & 0 & 0 & 1 & 1 & 0 & 0 & 0 & 0 & 0 & 0 & 1 & 0 & 0 & 0 & 0 
    \end{array}
\end{bmatrix}  \,.
\end{equation*}
According to the general construction, each row is characterized to have at most one non-vanishing {\blau entry} in each of the  four blocks and the number of rows corresponds to the number of measurements for any group $G=[1],[2],[3],[4]$.  

The  overall  CS matrix  acting on the 64 sensors  arranged in four  groups takes the  block diagonal form   
\begin{equation} \label{eq:CSmatrix} 
\Ao = \begin{bmatrix}
    \Ao_1 & 0 & 0 & 0 \\
    0 & \Ao_2 & 0 & 0 \\
    0 & 0 & \Ao_3 & 0 \\
    0 & 0 & 0 & \Ao_4 \\
\end{bmatrix} \in \{ 0,1 \}^{4 m_0 \times  64} \,,
\end{equation}
where $\Ao_G \in \{ 0,1\}^{m_0 \times 16}$ has the structure as in \eqref{eq:submatrix}. {\blau For these types of CS measurements combined with the sparsity paradigm, we address both the unique recovery question and the optimal design question. All matrices of the form \eqref{eq:submatrix} and \eqref{eq:CSmatrix} are experimentally implementable with the CS-PAPI system.}

{\blau
\begin{remark}[Selection of block size and group size]  The parameters guiding the types of CS measurements are the block size (sensors having the same switch) and the number of blocks per group. The product of these numbers gives the group size. The specific choices are determined by the current PAPI setting (block size four and four groups per block); however, they can be adjusted according to different experimental designs. For example, by fixing the group size to 16, another choice is a block size of two and eight groups per block. Such measurements are found to improve CS capabilities. However, on the downside, this doubles the number of fiber phase switches. Our framework is completely flexible in terms of group number and block size. The concrete choice should be determined by practical considerations.
\end{remark}
}

\subsection{Experimental realization}

\begin{figure}[htb!]
\centering
\includegraphics[width=0.8\textwidth]{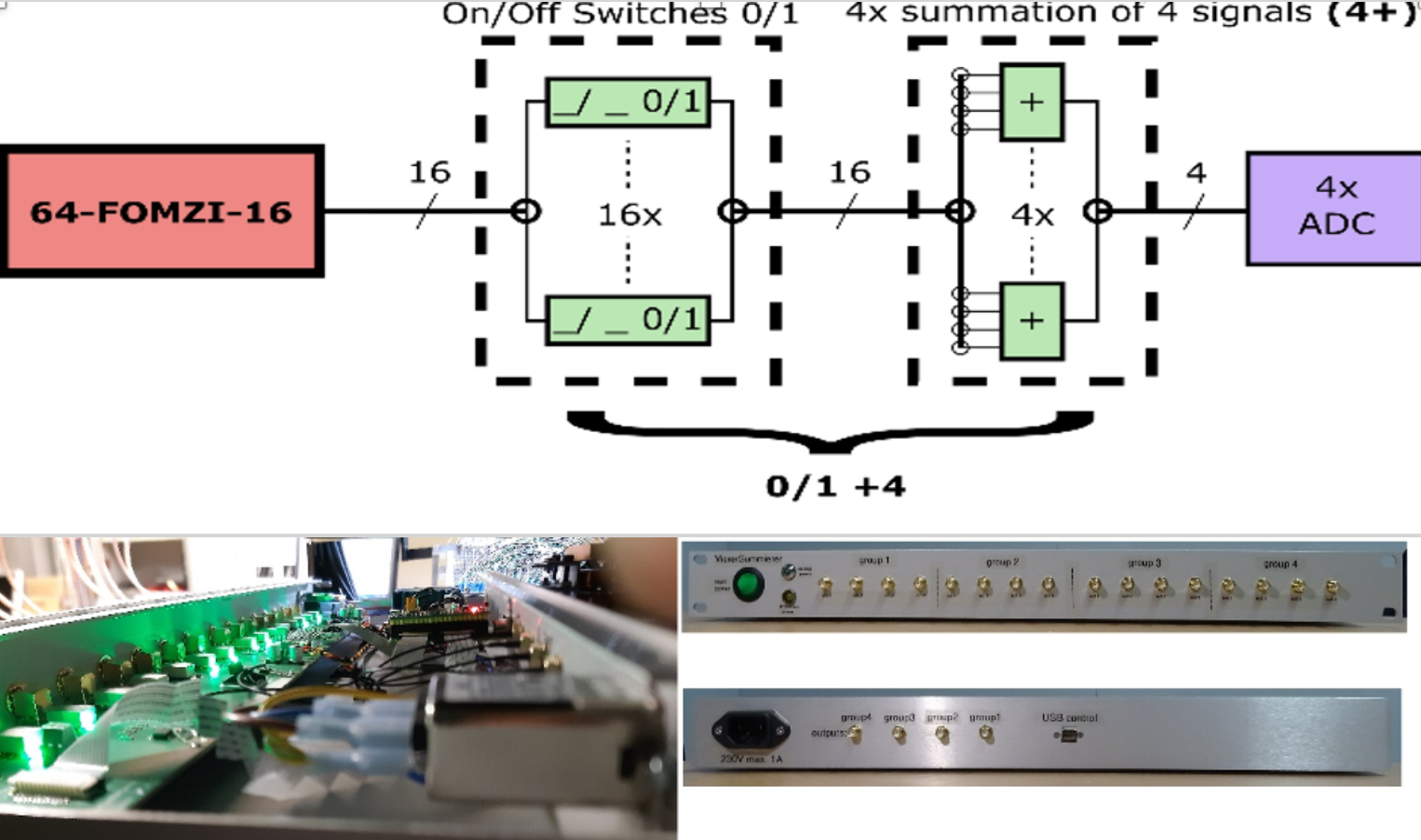}
\caption{Top: Illustration of the CS detection principle and its technical realization by SUM4. Bottom: Photo of the CS-PAPI system. {\blau Note that the CS module is located in the lower right part of Figure~\ref{fig:fomzi}, before the A/D converter. Additionally, the 4:1 switches are modified to allow for variable ILD selection.}}
\label{fig:cs-setup1}
\end{figure}

In order to technically implement CS on PAPI, a plug-and-play concept was developed by designing and implementing a CS module named SUM4 (for summing over 4) that can be integrated into the PAPI system. Recall {\blau that} before AD conversion, PAPI has 16 acoustic signals, where each signal corresponds to the ILD selected in the 16 blocks by the switch.
As a first step, we extend PAPI by enabling the arbitrary selection of ILDs within each group.  {\blau Additionally, we construct SUM4, where signals from four neighboring blocks are summed, resulting in four electrical signals that are sampled by the NI card.} Before summation each signal can be potentially {\blau be switched off,}  resulting in CS measurements of the form \eqref{eq:submatrix}, \eqref{eq:CSmatrix}. Figure~\ref{fig:cs-setup1} illustrates the schematic concept of SUM4, consisting of on/off switches, summation over blocks of four, and transmission to the ADC. Additionally, Figure~\ref{fig:cs-setup1} shows a photo of parts of the CS-PAPI system.

SUM4 can be seen as a device for analog signal conditioning and implements the CS aspects in the analog electrical domain. It allows the arbitrary superposition of up to four analog signals by switchable addition of the input signals. Additionally, the design permits the compensation of system-related losses in signal amplitudes, such as those caused by impedance matching. The low-noise design of the analog signal paths results in a signal-to-noise ratio of 80 dBV, corresponding to a resolution of at least 13 bits. The selection of electrical signals to be superimposed is done via the USB port. This involves implementing a virtual COM port with a custom control protocol. This intuitive control protocol facilitates easy integration of the device into a larger network of instruments via USB. To ensure optimal integration with PAPI, the quad summers combine four separate summing groups in one device, allowing 16 input signals to be routed to four independent outputs.

\subsection{Optimal design}

The CS-PAPI system with SUM4 allows us to perform any CS measurements  of the form \eqref{eq:submatrix}, \eqref{eq:CSmatrix}.   The aim in this section is to present a strategy for selecting  optimal  measurements within this class based on exact reconstruction. For that  purpose, we first note that  the measurements between the subgroups  are independent and thus we aim for optimal design of each $m_0 \times 16$ sub-matrix of the form \eqref{eq:submatrix}. Second, we focus on optimal design in the context of sparse recovery.  Thus we aim for binary  matrices $\Mo \in \R^{m_0 \times 16}$ of the form \eqref{eq:submatrix} with  $m_0 < 16$ {\blau which  allow us  to recover} sparse signals  $\signal \in \R^{16 \times 1}$  from data  $\Mo \signal$. Because the signal size is small, selecting these matrices at random  (as in standard CS) resulted in matrices not enabling  sparse recovery. We therefore  designed a quality measure and a strategy to construct matrices enabling sparse recovery.

A minimal requirement for the identifiability of sparse elements $\signal \in \R^{16}$ is the injectivity of $\Mo$ over the set of $s$-sparse elements. However injectivity alone is not sufficient in the sense that $\Mo \signal_1$ and $\Mo \signal_2$ can get close to each other for sparse signals $\signal_1, \signal_2$ very different from each other. Thus we actually need  to bound the difference $\norm{\Mo \signal_1 - \Mo  \signal_2}_2 $ in order to sufficiently separate $\signal_1$ and  $\signal_2$.  While this is essentially also included in the  RIP constant, in this paper we introduce a different concept which we think fits better to our aims. 

\begin{definition}[Sparse injectivity number]
For a matrix $\Mo \in \R^{m_0 \times n_0}$ and any  $s$ we define the $s$-sparse injectivity number ($s$-SIN) of $\Mo$ as
\begin{equation} \label{eq:SIB}
\spin_s (\Mo) \coloneqq \inf \Bigl\{ \frac{\norm{\Mo   \signal_1 - \Mo  \signal_2  }_2 }{  \norm{  \signal_1 -    \signal_2 }_2 }  \; \Big|  \; \signal_1   \neq \signal_2  \in \R^{n_0} \text{ are  $s$-sparse }  \Bigr\} \,.
\end{equation}
Alternatively, the $s$-SIN can be defined as the largest  constant $\spin \geq 0$ such that $\norm{  \Mo \signal_1 -  \Mo \signal_2  }_2 \geq  \spin \norm{ \signal_1 - \signal_2}_2 $ for  all $s$-sparse signals $\signal_1, \signal_2  \in \R^{n_0}$. 
\end{definition}

{\blau The $s$-SIN is strictly positive if and only if the matrix $\Mo$ is injective on the set of all $s$-sparse elements.} Unlike the usual RIP, it only asks for the one-sided estimate {\blau $\norm{ \Mo \signal_1 - \Mo \signal_2 }_2 \geq \spin \norm{ \signal_1 - \signal_2}_2 $. Furthermore, for $s \leq n_0/2$, it is easy to verify that $\sigma_s$ is the smallest singular value among all $m_0 \times 2s$ sub-matrices of $\mathbf{M}_0$.} 

A good CS matrix is a CS matrix with $\spin_s (\Mo)$ large relative to $\norm{\Mo}$. {\blau Values of $\spin_s (\Mo)$ greater than 0.1 have been empirically observed to result in stable and robust signal reconstruction.}  Randomly selecting $\Mo$ from our class of matrices turned out to very often yield (almost) vanishing $s$-SIN. On the other hand, computing the $s$-SIN for all admissible matrices to make an optimal selection is computationally infeasible. Therefore, to determine a suitable CS matrix, we use a simple algorithm where we repeatedly randomly select $\Mo$ from our CS matrix class and update the matrix whenever the {\blau $s$-SIN} is increased. This procedure is summarized in Algorithm~\ref{alg:selection}, where for PAPI we have $n_0 = 16$.

\begin{algorithm}
\caption{Optimized detector selection for CS matrix with large $s$-SIM} \label{alg:selection}
\begin{algorithmic}[1]
\State $\texttt{SINopt} \gets 0$ 
\State $\texttt{LISTopt}  \gets \texttt{zeros}(1,4)$
\State $\texttt{Mopt}  \gets \texttt{zeros}(m_0, n_0)$

\For{$i$ in $[1, \texttt{Niter}]$}
    \State $\texttt{LIST} \gets \texttt{random.sample}(m_0, n_0)$ \Comment{Draw  $m_0$ lists of active sensors}
    \State $\Mo \gets \texttt{makeCSMatrix}(\texttt{LIST})$  \Comment{Build the CS matrix}
    \State $\texttt{SIN} \gets \texttt{getSIN}(\Mo,s)$   \Comment{Compute the $s$-SIN of $\Mo$}

    \If{$\texttt{SIN} > \texttt{SINopt}$}
        \State $\texttt{LISTopt} \gets \texttt{LIST}$
        \State $\texttt{Mopt} \gets \Mo$
        \State $\texttt{SINopt} \gets \texttt{SIN}$
    \EndIf
\EndFor

\Return  \texttt{SINopt}, \texttt{LISTopt}, \texttt{Mopt} 
\Comment{Return optimal CS list, matrix and SIN}
\end{algorithmic}
\end{algorithm}

In Algorithm~\ref{alg:selection}, the function \texttt{random.sample} selects a feasible list of sensors and the function \texttt{makeCSMatrix} forms the corresponding CS matrix. Furthermore, \texttt{getSIN} computes the $s$-SIN. We have found empirically that procedure  results in  CS matrices with a SIN over $0.1$ in a reasonable time. Specifically, we take $m_0 = 12$ and $s=2$ for the results shown below.

{\blau Algorithm~\ref{alg:selection} can be extended to use block sizes other than four and numbers of blocks other than four. The only limiting factor is the increasing numerical complexity with increasing dimensions.}

\subsection{Two-step CS image reconstruction}

Due to the separable  nature of the image reconstruction problem \eqref{eq:ip} there  are naturally two  types of reconstruction methods, namely  one-step image reconstruction and two-step image  reconstruction. In the two-step methods,   the complete data  $ \Wo  \source $ are first recovered from CS data $\Ao [\Wo  \source] $ via  iterative methods, and in a second step  $\source$ is recovered from $ \Wo \source$ via wave inversion such as the FBP inversion formula. In the one-step approach, the initial pressure is directly  recovered from CS data using  iterative methods applied with the  full forward operator $\Ao \Wo$.  Both classes of methods come with certain strengths and limitations.  The two-step approach is fast as iterative signal reconstruction, is separated from the computationally costly evaluation of $\Wo$ and its adjoint.   Moreover, CS properties of  the matrix $\Ao$ can be exploited together with the sparsity  of $\Wo \source$, potentially after suitable basis transform. On the downside, image structure of $\source$ cannot be directly  integrated in the image reconstruction. {\blau The one-step approach, on the other hand, allows for easy integration of prior information about the image to be generated.} However, CS reconstruction theory  based on sparsity and specific properties of he forward matrix can hardly be integrated. Hybrid methods such as those proposed in \cite{ebner2022convergence} might overcome such issues. {\blau Another drawback of one-step approaches is that they necessitate the repeated use of the time-consuming evaluation of $\Wo$ and its adjoint.}

Due  to its clear interpretability and computational efficiency in this study we work with the two-step approach. {\blau Specifically, we utilize temporal transforms in combination with 1D total variation (TV) minimization.} For that purpose we  apply a transform $\To \colon \R^q \to \R^q$ acting in the time domain such that the transformed pressure $ P \To^T   $ has sparse gradients. Thus an approximation $H =[h_1^T, \dots, h_n^T]^T$ to $ P \To^T $ can be recovered by TV minimization
\begin{equation} \label{eq:tv}
 \norm{ \Ao H  - \csdata   \To^T}^2 + \norm{  \partial_1  H }_1 
 =
\sum_{\ell=1}^q  \norm{ \Ao h_\ell  - (\csdata   \To^T)_\ell}^2 + \norm{  \partial_1  h_\ell }_1    \to \min_H \,,
\end{equation}     
where $\partial_1$ is the derivative in the sensor direction.  Problem~\eqref{eq:tv}  can be solved by a series of 1D TV minimization problems for the 1D signals $h_\ell$ and is numerically efficient. Further, by writing the FBP formula \eqref{eq:fbp2d} as    
\begin{equation} \label{eq:fbp2d-mod}
          \source(\rr)
         =
        - \frac{1}{\pi R}
        \int_{\partial B_R}
        \int_{\abs{\rr-z}}^\infty
        \frac{ (\partial_t t   [\To^{-1} \circ \To \circ \Wo  \source])(\rss, t)}{ \sqrt{t^2-\sabs{\rr-\rss}^2}}  \, \rmd t
        \rmd S(\rss)
\end{equation}
we can recover the unknown $\source$ from the filtered signals $\To \circ \Wo  \source$ in the first step. Equation  \eqref{eq:tv} and \eqref{eq:fbp2d-mod} constitute the two-step method  we  use for image reconstruction in this paper

\begin{remark}
Let us mention some further work on image reconstruction in CSPAT. Using intertwining relations between spatial and temporal operations for the wave equation, we extended the sparsifying transform approach to the image domain \cite{haltmeier2018sparsification, zangerl2021multiscale}, enabling one-step inversion. This and the two-step method can also be applied to CSPAT with standard point-like measurements. Other early work on CSPAT has been done in \cite{provost2009application, guo2010compressed, alberti2021compressed, arridge2016accelerated, huynh2019single}, where various compressive sampling strategies have been used with sparse recovery techniques. Recently, machine learning methods have been used in the context of CSPAT \cite{hauptmann2018model, antholzer2019nett, grohl2021deep, hauptmann2020deep, anastasio2023deep, davoudi2019deep, antholzer2019deep}.
\end{remark}

{\blau 
\section{Numerical experiments}

Due to the restricted CS matrices, it is challenging to achieve even a small compression factor $n/m$. Note that for our structured CS matrices we require sparsity within the 4 groups of 16 sensors each. For the following numerical investigation, we use a sparsity level of $s=2$. Numerically, it turns out that we need 12 measurements to obtain a non-singular SIN with the algorithm outlined above.

\begin{figure}[htb!]
\centering \includegraphics[width=0.8\textwidth]{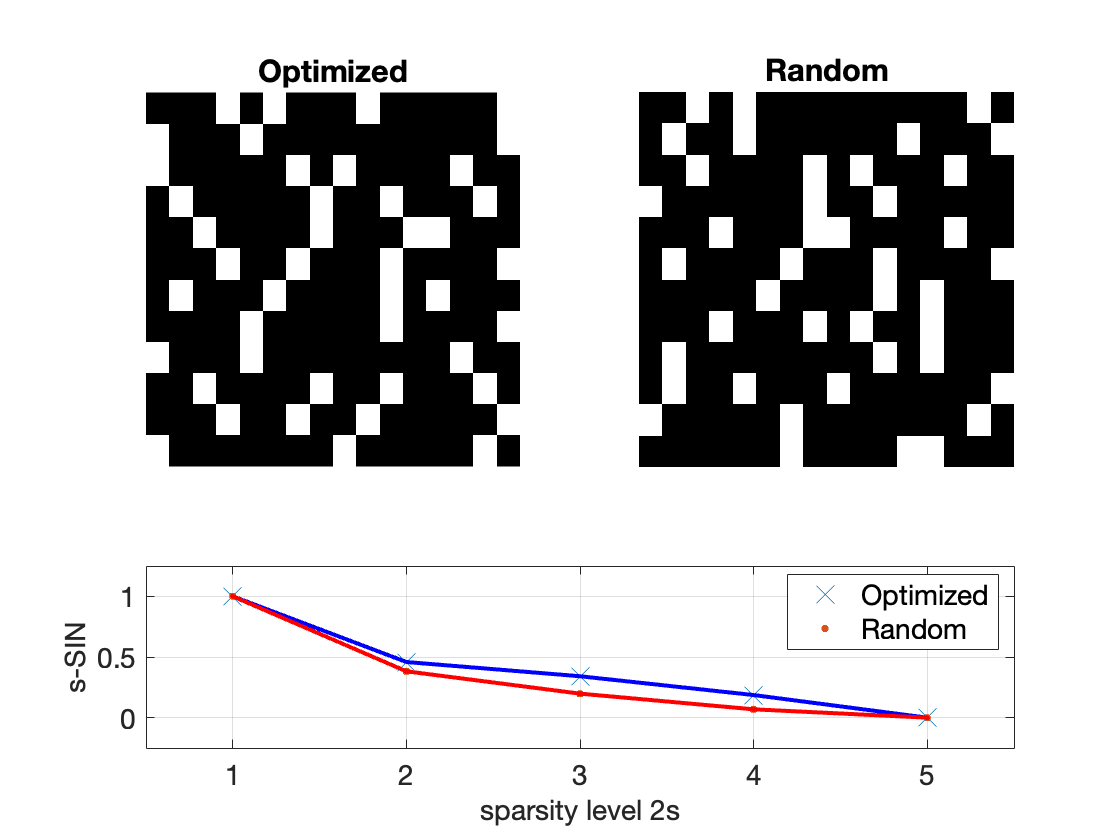}
\caption{\blau Comparison of a measurement matrix for optimized $2$-SIN (top left) and a random matrix according to the CS setup (top right), which has been corrected to have non-vanishing SIN. The bottom image shows the computed $s$-SIN for the sparsity level $2s = 1,2,3,4,5$.} \label{fig:CSmat}
\end{figure}

\subsection{Measurement design results}

We use the parameters of our PAPI system, where measurements on a group of detectors have the form \eqref{eq:submatrix}, whose structure is determined by the size of the blocks (which is 4 for our PAPI) and the number of blocks within a group (which is also 4 for our PAPI).   The goal of CS is to keep the number $m_0$ of measurements small, while allowing the unique recovery of certain elements.  Following the sparsity paradigm, our approach is to use algorithm~\ref{alg:selection} to find a matrix with non-singular $s$-SIN. The larger $s$, the more general the signal class, but the less likely it is to get a non-vanishing $s$-SIN. So we take $2s=4$ to have at least some generality in the signal class.  

Running Algorithm~\ref{alg:selection} we found that a non-singular SIN could be found for $m_0=12$ measurements. In particular, in almost every test run with 100 iterations, we could find a matrix with a SIN of about $0.14$, which we then selected.  Even for $m=11$ we could find such matrices after a longer search.  However, we could not increase the compression factor further in the sense that for $m=10$, even after 100000 iterations, no SIN larger than machine precision could be found. Roughly speaking, our work demonstrates a compression factor of at least $4/3$ for block size 4 and group size 16.

\begin{remark}[Variable block size and group size]
In order to put our work into a broader perspective, it is worth investigating whether different block sizes and numbers of blocks result in a larger compression factor. Testing our algorithm with the same group size but a block size of two, we found that indeed, using $m=10$ measurements results in a nonsingular $s$-SIN of approximately 0.21, demonstrating an increased compression factor of $8/5$. A similar effect has been observed when keeping the block size constant while increasing the group size.
\end{remark}

Having a non-singular $2$-SIR allows for theoretical exact recovery of $2$-sparse signals from exact data. In reality, robustness regarding noise and stability concerning the sparsity level using specific reconstruction algorithms are central. While this is not part of our theory, we expect similar results to the (unfortunately asymptotic) theory of CS. Our numerical results below support this.

\begin{figure}[htb!]
\centering
\includegraphics[width=\textwidth]{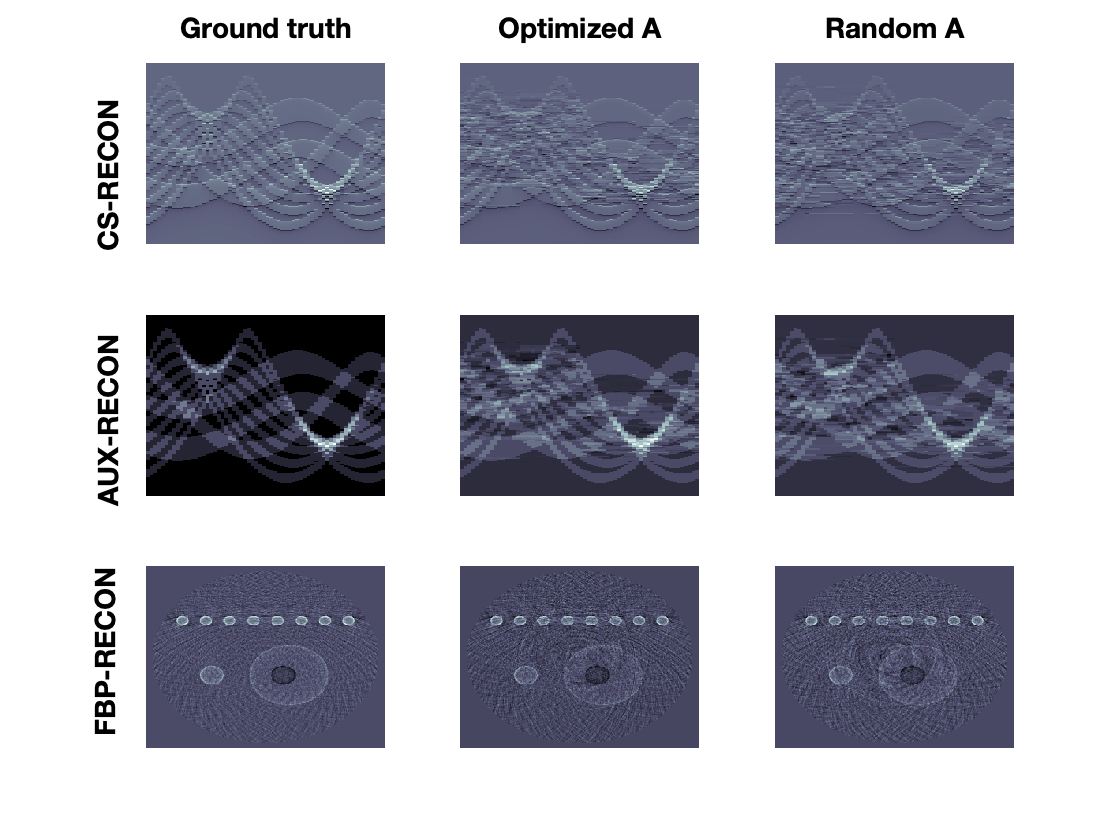}
\caption{Reconstruction results from exact data data. Top row (from left to right): Data from 64 ILD, reconstruction using an optimized CS matrix and  reconstruction using a random matrix.  Middle row: corresponding time-transformed data from the pressure. Bottom:  Corresponding FBB reconstructions. {\blau  In the first two rows, the horizontal direction represents the spatial dimension, while the vertical direction represents time.}}
\label{fig:rec-exact}
\end{figure}

\begin{figure}[htb!]
\centering
\includegraphics[width=\textwidth]{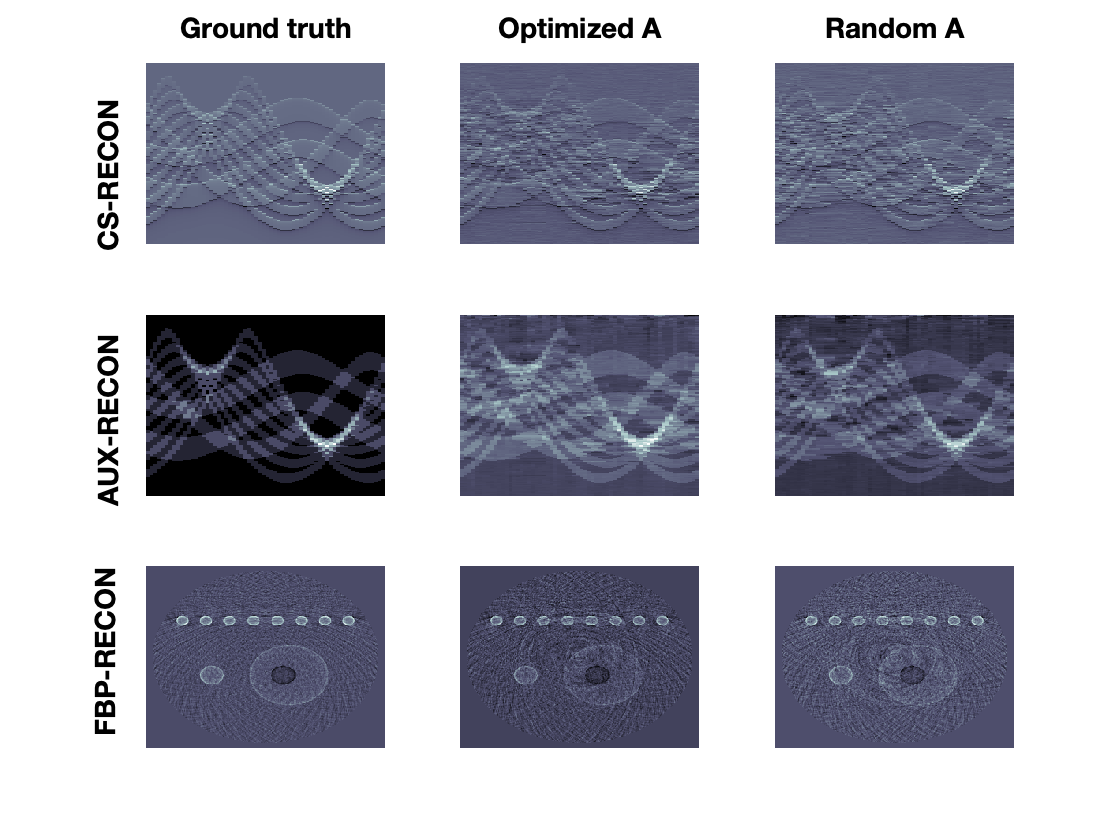}
\caption{Reconstruction results from noisy data. Top row (from left to right): Data from 64 ILD, reconstruction using an optimized CS matrix and  reconstruction using a random matrix.  Middle row: corresponding time-transformed data from the pressure. Bottom:  Corresponding FBB reconstructions. {\blau  In the first two rows, the horizontal direction represents the spatial dimension, while the vertical direction represents time.}}
\label{fig:rec-noisy}
\end{figure}

\begin{figure}[htb!]
\centering
\includegraphics[width=\textwidth]{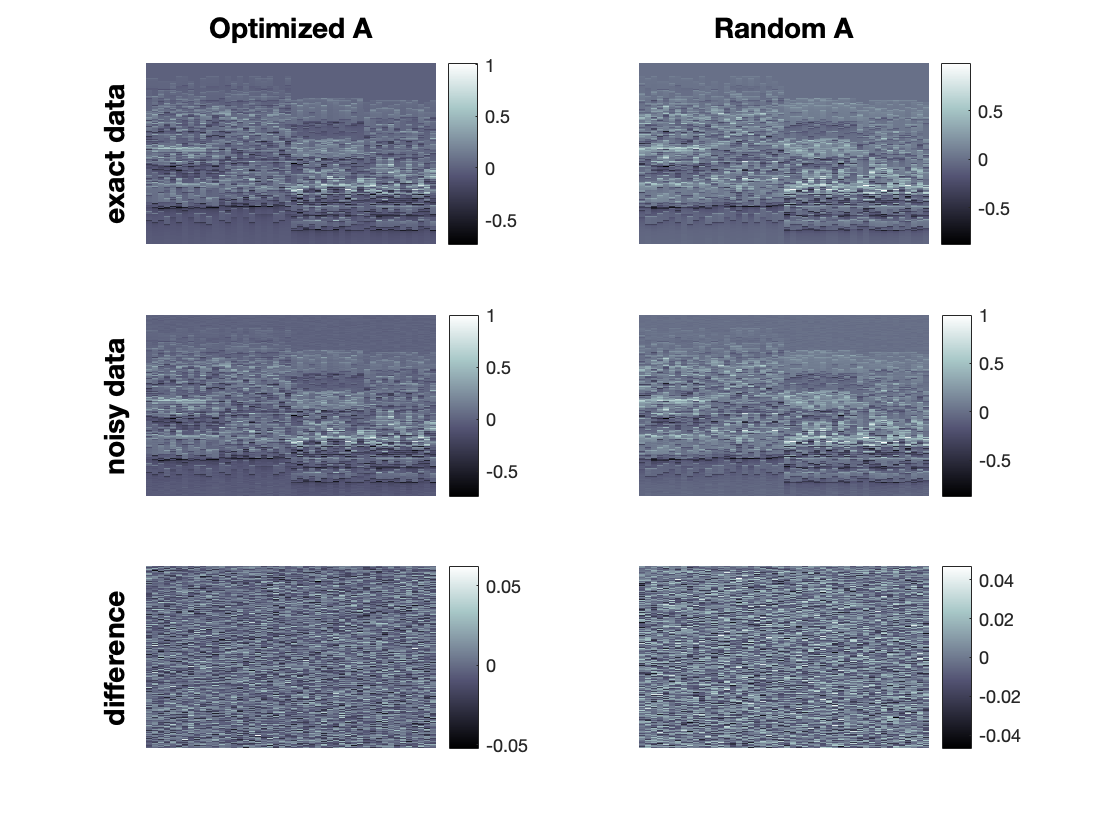}
\caption{Exact and noisy data for the optimized and the random matrix. {\blau  The  horizontal direction represents the spatial dimension, while the vertical direction represents time.} }
\label{fig:data}
\end{figure}
 
}

\begin{figure}[htb!]
\centering
\includegraphics[width=\textwidth]{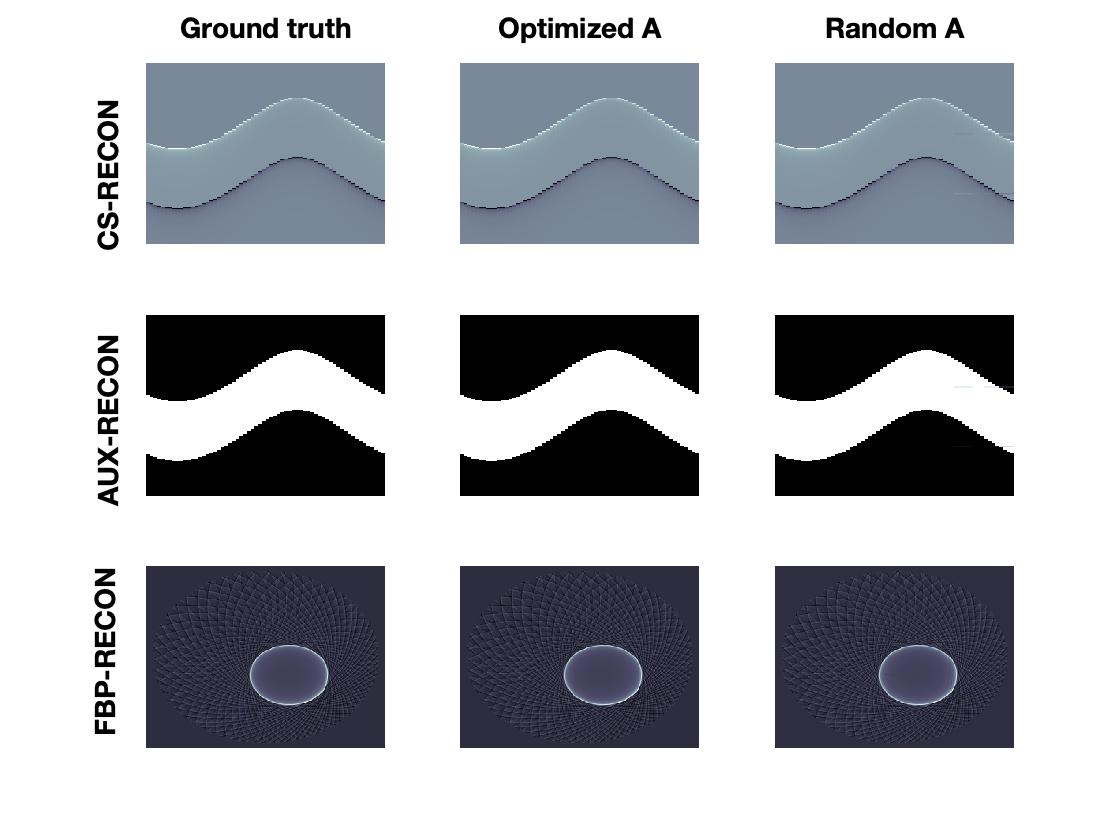}
\caption{Reconstruction results of sparse object. Top row (from left to right): Data from 64 ILD, reconstruction using an optimized CS matrix and  reconstruction using a random matrix.  Middle row: corresponding time-transformed data from the pressure. Bottom:  Corresponding FBB reconstructions. {\blau  In the first two rows, the horizontal direction represents the spatial dimension, while the vertical direction represents time.}}
\label{fig:rec-sparse}
\end{figure}

\subsection{Image reconstruction results}

For image reconstruction, we use the two-step sparse recovery method described above. The key there is to apply a temporal transform to obtain sparsity. Here we use a phantom such that the spherical means are piecewise constant. Thus, in the first step, we use the Abel transform as the time transform and recover the spherical means using TV minimization \eqref{eq:tv}.

Reconstruction results from exact and noisy data are shown in Figure \ref{fig:rec-exact}.  We use two different measurement matrices, the first one is found by our algorithm and the second one is a randomly selected matrix from the CSPAT family that we corrected by educated guess to get non-vanishing 2-SIN. The CS measurement data and the added noise are shown in Figure \ref{fig:data}. For specific parameter settings, we refer to the Matlab code that will be made publicly available.  We consider the FBP reconstruction as our ground truth because our aim  is to approximate the image quality achieved with the full sensor array (64 sensors). Our ground truth phantom consists of circles, but they are not homogeneous. The profile has been chosen such that the spherical means of the circular regions are piecewise constant, making it well-suited for total variation (TV) minimization. In this way, we avoid a transformation that modifies the signal in that regard, as suggested in \cite{sandbichler2015novel}.

We find that the reconstruction procedure is indeed very stable and robust. In particular, the noise had a small negative impact on the results.  The reconstruction artifacts are due to the failure of the strict $2$-sparsity assumption. To support such a claim, we also show results (Figure \ref{fig:rec-sparse}) for a simple phantom where $2$-sparsity on the 16-groups almost holds. In this case, the CS reconstruction hardly differs from the ground truth. For precise relative error values see Table \ref{tab:error}. All reconstruction results demonstrate stability and robustness.

%

%

%

    \begin{table}[htb!]
    \centering
    \caption{Relative. $\ell^2$ error in the data (row 1), the CS reconstruction  (row 2) and the final FBP reconstruction error using the optimized  and 
   random  matrix (row 3). }
    \begin{tabular}{|l|ccc|ccc|}
        \toprule
         \multicolumn{4}{|r}{optimized $A$} 
        & \multicolumn{3}{r|}{random $A$}
        \\
        \midrule
         & data  & CS   & FBP  & data & CS   & FBP  \\
        \midrule
        non-sparse phantom (noisy)  &  0.0901 & 0.3159 & 0.6686 & 0.0766 & 0.3414 & 0.6810   \\
        non-sparse phantom (exact) &  x & 0.2594 & 0.6049 & x & 0.2849 & 0.6571  \\
        sparse phantom (exact) &  x & 0.0003 & 0.0010 & x & 0.0107 & 0.0509\\
        \bottomrule
    \end{tabular} \label{tab:error}
\end{table}

\section{Conclusion and outlook}
\label{sec:concluson}

In this paper we presented the experimental realization of a CS-PAPI system extending the existing tomograph. We demonstrated that the specific setup allows perfect recovery  of sparse signals. However for that purpose we could not select an admissible matrix uniformly at random, but a systematic strategy exploiting the SIM.

One future task is to go beyond the sparsity model. Thus our aim is to find  CS matrices $\Ao \in \R^{m \times 16}$ not targeting sparsity but actual real data. This can be done two-fold. First one can train a matrix such that  $16 \times 1$ pieces in data domain  are optimally separated. Second optimization can be improved by optimizing over the image space. {\blau  This allows us to consider that, due to the forward map $\Wo$, the $16 \times 1$ patches are actually correlated since they originate from the same initial source.} Deep  learning  and neural networks are natural candidates unveiling such hidden correlation.

\section*{Disclosures}
None of the authors have any potential conflicts of interest to disclose.

\section*{Code, Data, and Materials Availability} 
 
The code for generating a CS matrix with large SIN as well for producing the shown numerical results are provided upon request. No additional  data  is required for this study.

\section*{Acknowledgment}
This work has been supported by the Austrian Science Fund (FWF), projects P 30747-N32 and P 33019-N.

\end{document}